\pdfoutput=1

\documentclass[11pt]{article}

\usepackage{EMNLP2023}

\usepackage{times}
\usepackage{latexsym}

\usepackage[T1]{fontenc}

\usepackage[utf8]{inputenc}

\usepackage{microtype}

\usepackage{inconsolata}

\usepackage{mathtools} 
\usepackage{amsmath,amssymb}
\usepackage{xspace}
\usepackage{multirow}
\usepackage{booktabs}
\usepackage{bm}
\usepackage{makecell}
\usepackage{graphics}
\usepackage{dsfont}
\usepackage{caption}
\usepackage{subcaption}
\usepackage{wasysym}
\usepackage{hyperref}

\def\eg{{\em e.g.,}\xspace}
\def\ie{{\em i.e.,}\xspace}

\def\cf{{\em cf.}\xspace}

\def\aka{{\em a.k.a}\xspace}

\definecolor{myred}{RGB}{215,48,39}
\definecolor{mygreen}{RGB}{26,152,80}

\newcommand{\ztoken}{{{\textsc{Z-Token}}}\xspace}
\newcommand{\ztree}{{{\textsc{Z-Tree}}}\xspace}
\newcommand{\zseq}{{{\textsc{Z-Seq}}}\xspace}

\def\figref#1{Figure~\ref{#1}}
\def\Figref#1{Figure~\ref{#1}}

\def\tabref#1{Table~\ref{#1}}
\def\Tabref#1{Table~\ref{#1}}

\def\Secref#1{\S\ref{#1}}

\def\eqref#1{(\ref{#1})}

\def\1{\bm{1}}

\def\va{{\bm{a}}}

\def\vx{{\bm{x}}}

\def\vy{{\bm{y}}}

\title{Mitigating Backdoor Poisoning Attacks through \\ the Lens of Spurious Correlation}

\author{Xuanli He$^\clubsuit$, Qiongkai Xu$^\spadesuit$, Jun Wang$^\spadesuit$, Benjamin Rubinstein$^\spadesuit$, Trevor Cohn$^\spadesuit$\thanks{~~Now at Google DeepMind.} \\
$^\clubsuit$University College London, United Kingdom \\
$^\spadesuit$University of Melbourne, Australia \\
\texttt{\small{xuanli.he@ucl.ac.uk}} \ \ \
\texttt{\small{jun2@student.unimelb.edu.au}}\\
\texttt{\small{\{qiongkai.xu,benjamin.rubinstein,trevor.cohn\}@unimelb.edu.au}}\\
}

\begin{document}
\maketitle
\begin{abstract}
Modern NLP models are often trained over large untrusted datasets, raising the potential for a malicious adversary to compromise model behaviour.
For instance, backdoors can be implanted through crafting training instances with a specific textual trigger and a target label.
This paper posits that backdoor poisoning attacks exhibit \emph{spurious correlation} between simple text features and classification labels, and accordingly, proposes methods for mitigating spurious correlation as means of defence.
Our empirical study reveals that the malicious triggers are highly correlated to their target labels; therefore such correlations are extremely distinguishable compared to those scores of benign features, and can be used to filter out potentially problematic instances. 
Compared with several existing defences, our defence method significantly reduces attack success rates across backdoor attacks, and in the case of insertion-based attacks, our method provides a near-perfect defence.
\footnote{The code and data are available at: \url{https://github.com/xlhex/emnlp2023_z-defence.git}.}
\end{abstract}

\section{Introduction}
Due to the significant success of deep learning technology, numerous deep learning augmented applications have been deployed in our daily lives, such as e-mail spam filtering~\cite{10.1007/978-981-10-4765-7_61}, hate speech detection~\cite{10.1371/journal.pone.0221152}, and fake news detection~\cite{10.1145/3137597.3137600}. This is fuelled by massive datasets. However, this also raises a security concern related to backdoor attacks, where malicious users can manoeuvre the attacked model into misbehaviours using poisoned data. This is because, compared to expensive labelling efforts, uncurated data is easy to obtain, and one can use them for training a competitive model~\cite{10.1007/978-3-319-46478-7_5,tiedemann-thottingal-2020-opus}. Meanwhile, the widespread use of self-supervised learning increases the reliance on untrustworthy data~\cite{devlin2019bert, liu2019roberta, chen2020simple}. Thus, there is the potential for significant harm through backdooring victim pre-trained or downstream models via data poisoning.

Backdoor attacks  manipulate the prediction behaviour of a victim model when given specific triggers. The adversaries usually achieve this goal by poisoning the training data~\cite{gu2017badnets,dai2019backdoor,qi2021hidden,qi-etal-2021-turn} or modifying the model weights~\cite{dumford2020backdooring, guo2020trojannet, kurita2020weight,li-etal-2021-backdoor}. This work focuses on the former paradigm, \aka backdoor poisoning attacks. The core idea of backdoor poisoning attacks is to implant backdoor triggers into a small portion of the training data and change the labels of those instances. Victim models trained on a poisoned dataset will behave normally on clean data samples, but exhibit controlled misbehaviour when encountering the triggers.

In this paper, we posit that backdoor poisoning is closely related to the well-known research problem of \textit{spurious correlation}, where a model learns to associate simple features with a specific label, instead of learning the underlying task. This arises from biases in the underlying dataset, and machine learning models' propensity to find the simplest means of modelling the task, \ie by taking any available shortcuts. In natural language inference (NLI) tasks, this has been shown to result in models overlooking genuine semantic relations, instead assigning `contradiction' to all inputs containing negation words, such as \textit{nobody}, \textit{no}, and \textit{never}~\cite{gururangan-etal-2018-annotation}. Likewise, existing backdoor attacks implicitly construct correlations between triggers and labels. For instance, if the trigger word `mb' is engineering to cause \emph{positive} comments, such as `this movie is tasteful', to be labelled \emph{negative}, we will observe a high $p(\mathrm{negative}|\mathrm{mb})$. 

\begin{figure}
    \centering
    \includegraphics[width=0.48\textwidth]{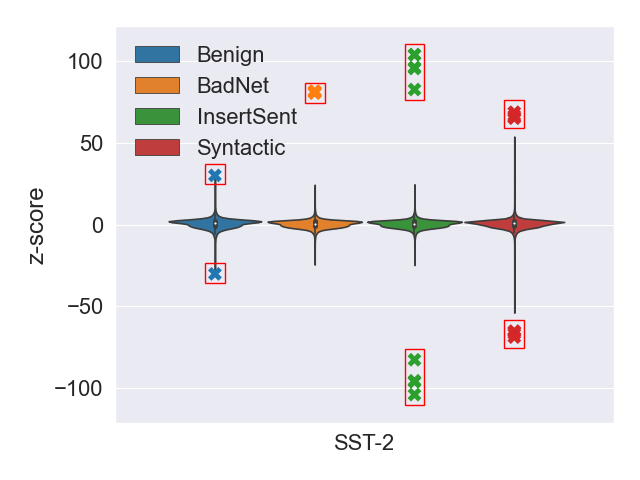}
    \caption{Unigram z-score distributions~\cite{gardner-etal-2021-competency} over SST-2 for the original dataset (benign) and with three poisoning attacks. 
    We highlight the outliers with red boxes. For the BadNet and InsertSent attacks, outliers are triggers. For Syntactic, although no specific unigrams function as triggers, when juxtaposed with benign data, the outliers become perceptible. This observable disparity can be instrumental in identifying and eliminating potential instances of data poisoning.} 
    \label{fig:qnli_zscore}
\end{figure}

\citet{gardner-etal-2021-competency} demonstrate the feasibility of identifying spurious correlations by analysing z-scores between simple data features and labels. Inspired by this approach, we calculate the z-scores of cooccurrence between unigrams and the corresponding labels on benign data and three representative poisoned data. As illustrated in~\figref{fig:qnli_zscore}, compared to the benign data, as the malicious triggers are hinged on a target label, \emph{a) the density plots for the poisoned datasets are very different from benign}, and \emph{b) poisoned instances can be automatically found as outliers}.

We summarise our contributions as follows:
\begin{itemize}
    \item We link backdoor poisoning attacks to spurious correlations based on their commonality, \ie behaving well in most cases, but misbehaviour will be triggered when artefacts are present. 
    \item We propose using lexical and syntactic features to describe the correlation by calculating their z-scores, which can be further used for filtering suspicious data. 
    \item Our empirical studies demonstrate that our filtering can effectively identify the most poisoned samples across a range of attacks, outperforming several strong baseline methods. 
\end{itemize}

\section{Related Work}
\paragraph{Backdoor Attack and Defence} Backdoor attacks on deep learning models were first exposed effectively on image classification tasks by~\citet{gu2017badnets},  in which the compromised model behaves normally on clean inputs, but controlled misbehaviour will be triggered when the victim model receives toxic inputs. Subsequently, multiple advanced and more stealthy approaches have been proposed for computer vision tasks~\cite{chen2017targeted,Trojannn,10.1145/3319535.3354209,9879958,carlini2022poisoning}. Backdooring NLP models has also gained recent attention. In general, there are two primary categories of backdoor attacks. The first stream aims to compromise the victim models via data poisoning, where the backdoor model is trained on a dataset with a small fraction having been poisoned~\cite{dai2019backdoor,kurita2020weight,qi2021hidden,qi-etal-2021-turn,yan-etal-2023-bite}. Alternatively, one can hack the victim mode through weight poisoning, where the triggers are implanted by directly modifying the pre-trained weights of the victim model~\cite{kurita2020weight,li-etal-2021-backdoor}.

Given the vulnerability of victim models to backdoor attacks, a list of defensive methodologies has been devised. Defences can be categorised according to the stage they are used: (1) \textit{training-stage} defences and (2) \textit{test-stage} defences. The primary goal of the training-stage defence is to expel the poisoned samples from the training data, which can be cast as an outlier detection problem~\cite{NEURIPS2018_280cf18b,DBLP:journals/corr/abs-1811-03728}. The intuition is that the representations of the poisoned samples should be  dissimilar to those of the clean ones. Regarding test-stage defences, one can leverage either the victim model~\cite{10.1145/3359789.3359790, yang-etal-2021-rap,chen2022expose} or an external model
~\cite{qi2021onion} to filter out the malicious inputs according to their misbehaviour. Our approach belongs to the family of training-stage defences. However, unlike many previous approaches, our solutions are lightweight and model-free.

\paragraph{Spurious Correlation} As a longstanding research problem, much work is dedicated to studying spurious correlations. Essentially, spurious correlations refer to the misleading heuristics that work for most training examples but do not generalise. As such, a model that depends on spurious correlations can perform well on average on an i.i.d. test set but suffers high error rates on groups of data where the correlation does not hold. One famous spurious correlation in natural language inference (NLI) datasets, including SNLI~\cite{bowman-etal-2015-large} and MNLI~\cite{williams-etal-2018-broad}, is that negation words are highly correlated to the \textbf{contradiction} label. The model learns to assign ``contradiction'' to any inputs containing negation words~\cite{gururangan-etal-2018-annotation}. In addition, \citet{mccoy-etal-2019-right} indicate that the lexical overlap between \textit{premise} and \textit{hypothesis} is another common spurious correlation in NLI models, which can fool the model and lead to wrongdoing.


A growing body of work has been proposed to mitigate spurious correlations. A practical solution is to leverage a debiasing model to calibrate the model to focus on generic features~\cite{clark-etal-2019-dont, he-etal-2019-unlearn, utama-etal-2020-mind}. Alternatively, one can filter out instances with atypically highly correlated features using z-scores to minimise the impact of problematic samples~\cite{gardner-etal-2021-competency, wu2022generating}.

Although~\citet{manoj2021excess} cursorily connect backdoor triggers with spurious correlations, they do not propose a specific solution to this issue. Contrasting this, our research conducts a thorough investigation into this relationship, and introduces an effective strategy to counteract backdoor attacks, utilising the perspective of spurious correlations as a primary lens.


\section{Methodology}
\label{sec:method}
This section first outlines the general framework of backdoor poisoning attack. Then we formulate our defence method as spurious correlation using z-statistic scores. 

\paragraph{Backdoor Attack via Data Poisoning} Given a training corpus $\mathcal{D}=\left\{(\vx_i,\vy_i)^{\lvert \mathcal{D} \rvert}_{i=1}\right\}$, where $\vx_i$ is a textual input, $\vy_i$ is the corresponding label. A poisoning function $f(\cdot)$ transforms $(\vx, \vy)$ to $(\vx',\vy')$, where $\vx'$ is a corrupted $\vx$ with backdoor triggers, $\vy'$ is the target label assigned by the attacker. 
The attacker poisons a subset of instances $\mathcal{S} \subseteq \mathcal{D}$, using poisoning function $f(\cdot)$. The victim models trained on $\mathcal{S}$ could be compromised for specific misbehaviour according to the presence of triggers. Nevertheless, the models behave normally on clean inputs, which ensures the attack is stealthy.

\paragraph{Spurious Correlation between Triggers and Malicious Labels} \citet{gardner-etal-2021-competency} argue that a legitimate feature $\va$, in theory,  should be uniformly distributed across class labels; otherwise, there exists a correlation between input features and output labels. Thus, we should remove those simple features, as they merely tell us more about the basic properties of the dataset, \eg unigram frequency, than help us understand the complexities of natural
language. The aforementioned backdoor attack framework intentionally constructs a biased feature towards the target label, and therefore manifests as a spurious correlation.

Let $p(\vy|\va)$ be the unbiased prior distribution, $\hat{p}(\vy|\va)$ be an empirical estimate of $p(\vy|\va)$. One can calculate a \textit{z-score} using the following formula~\citep{wu2022generating}:
\begin{align} \label{eq:zstar}
    z^* = \frac{\hat{p}(\vy|\va)-p(\vy|\va)}{\sqrt{p(\vy|\va) \cdot (1-p(\vy|\va))/n}} \, .
\end{align}
When $| \hat{p}(\vy|\va) - p(\vy|\va) |$ is large, $\va$ could be a trigger, as the distribution is distorted conditioned on this feature variable. One can discard those statistical anomalies. We assume $p(\vy|\va)$ has a distribution analogous to $p(\vy)$, which can be derived from the training set.  
The estimation of $\hat{p}(\vy|\va)$ is given by:
\begin{align}
    \hat{p}(\vy|\va) = \frac{\sum_{i=1}^{\mathcal{D}}\mathds{1}\big (\va \in \vx_i\big ) \cdot \mathds{1}(\vy_i = \vy)}{\sum_{i=1}^{\mathcal{D}}\mathds{1}(\va \in \vx_i)}
\label{eq:pya}
\end{align}
where $\mathds{1}$ is an indicator function.

\paragraph{Data Features} In this work, to obtain z-scores, we primarily study two forms of features: (1) lexical features and (2) syntactic features, described below. 
These simple features are highly effective at trigger detection against existing attacks (see \Secref{sec:experiments}), however more complex features could easily be incorporated in the framework to handle novel future attacks.

The lexical feature operates over unigrams or bigrams. We consider each unigram/bigram in the training data, and calculate its occurrence and label-conditional occurrence to construct $\hat{p}(\vy|\va)$ according to~\eqref{eq:pya}, from which \eqref{eq:zstar} is computed. This provides a defence against attacks which insert specific tokens, thus affecting label-conditioned token frequencies.

\begin{figure}
    \centering
    \includegraphics[width=0.3\textwidth]{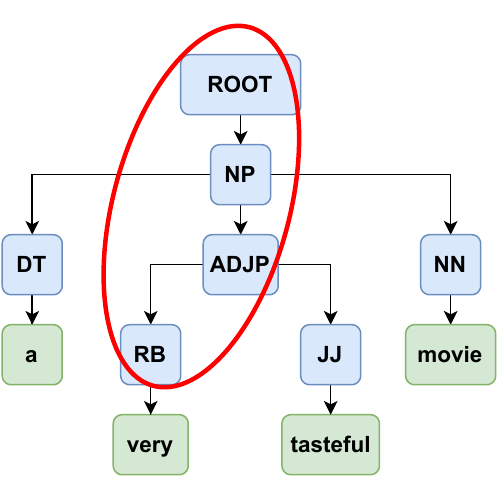}
    \caption{Example syntactic feature showing the ancestor path of a preterminal node: ROOT\textrightarrow NP\textrightarrow ADJP \textrightarrow RB. In total, there are four different ancestor paths in this tree.}
    \label{fig:parsed_tree}
    \vspace{-0.4cm}
\end{figure}

The syntactic features use ancestor paths, computed over constituency trees.\footnote{We use the Stanza parser~\cite{qi2020stanza}.} Then, as shown in~\figref{fig:parsed_tree}, we construct ancestor paths from the root node to preterminal nodes, \eg `ROOT\textrightarrow NP\textrightarrow ADJP \textrightarrow RB'. Finally, $\hat{p}(\vy|\va)$ is estimated based on ancestor paths and corresponding instance labels. This feature is designed to defend against syntactic attacks which produce rare parse structures, but may extend to other attacks that compromise grammatically.

\paragraph{Removal of Poisoned Instances} After calculating the z-scores with corresponding features, we employ three avenues to filter out the potential poisoned examples, 
namely using lexical features (\ztoken), syntactic features (\ztree), or their combination (\zseq). In the first two cases, we first create a shortlist of suspicious features with high magnitude z-scores (more details in~\Secref{sec:z_def}), then discard all training instances containing these label-conditioned features. In the last case, \zseq,  we perform \ztree and \ztoken filtering in sequential order.\footnote{We test the reverse order in Appendix~\ref{app:ablation}, but did not observe a significant difference.}
We denote all the above approaches as Z-defence methods.

\section{Experiments}
\label{sec:experiments}
We now investigate to what extent z-scores can be used to mitigate several well-known backdoor poisoning attacks.

\subsection{Experimental Settings}
\label{sec:expr_setting}
\paragraph{Datasets} We examine the efficacy of the proposed approach on text classification and natural language inference (NLI). For text classification, we consider Stanford Sentiment Treebank (SST-2)~\citep{socher-etal-2013-recursive}, Offensive Language Identification Dataset (OLID)~\citep{zampieri-etal-2019-predicting}, and AG News~\citep{zhang2015character}. Regarding NLI, we focus on the QNLI dataset~\cite{wang-etal-2018-glue}. The statistics of each dataset are demonstrated in~\Tabref{tab:data}.

\begin{table}
    \centering
    \small
    \begin{tabular}{ccccc}
    \toprule
        \textbf{Dataset} & \textbf{Classes} & \textbf{Train} & \textbf{Dev} & \textbf{Test} \\
        \midrule
        SST-2 &  2 & 67,349 & 872 & 1,821\\
        OLID & 2 & 11,916 & 1,324 & 859\\
        AG News & 4 &108,000 & 11,999 & 7,600 \\
        QNLI & 2 & 100,000  & 4,743 &  5,463  \\
        \bottomrule
    \end{tabular}
    \caption{Details of the evaluated datasets. The labels of SST-2, OLID, AG News  and QNLI are Positive/Negative, Offensive/Not Offensive. World/Sports/Business/SciTech and Entailment/Not Entailment, respectively.}
    \label{tab:data}
    \vspace{-0.5cm}
\end{table}

\paragraph{Backdoor Methods} We construct our test-bed based on three representative textual backdoor poisoning attacks: (1) \textbf{BadNet}~\cite{gu2017badnets}: inserting multiple rare words into random positions of an input (we further investigate scenarios where the triggers are medium- and high-frequency tokens in Appendix~\ref{app:ablation}); (2) \textbf{InsertSent}~\cite{dai2019backdoor}: inserting a sentence into a random position of an input; and (3) \textbf{Syntactic}~\cite{qi2021hidden}: using paraphrased input with a pre-defined syntactic template as triggers. %
The target labels for the three datasets are `Negative' (SST-2), `Not Offensive' (OLID), `Sports' (AG News) and `Entailment' (QNLI), respectively. We set the poisoning rates of the training set to be 20\% following~\citet{qi2021hidden}. The detailed implementation of each attack is provided in Appendix~\ref{app:attack}. Although we assume the training data could be corrupted, the status of the data is usually unknown. Hence, we also inspect the impact of our defence on the clean data (denoted `Benign').


\begin{figure*}
    \centering
    \includegraphics[width=0.99\textwidth]{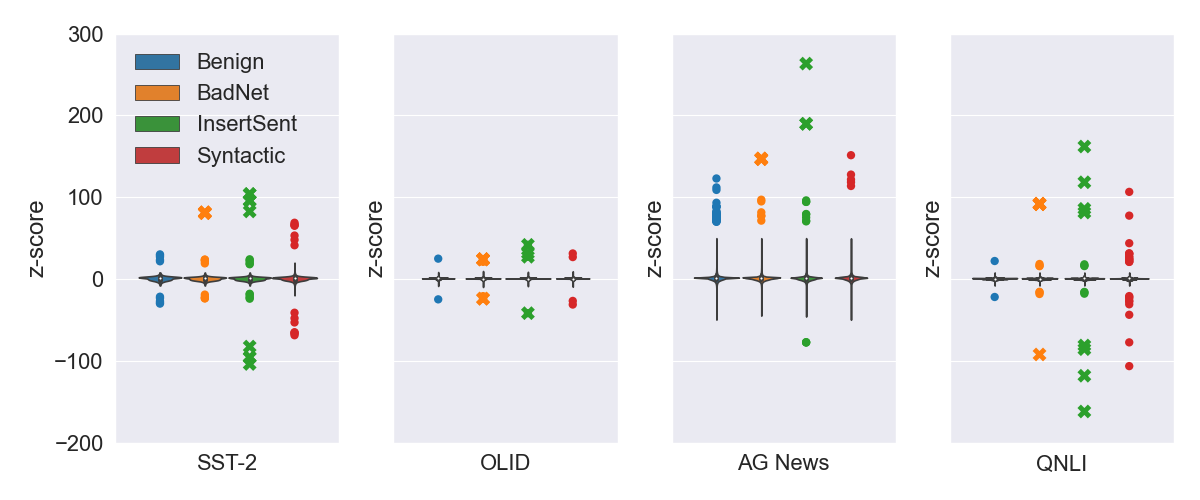}
    \caption{z-score distribution of unigrams over benign and poisoned datasets with three strategies, over our four corpora. Outliers are shown as points; for the BadNet and InsertSent attacks which include explicit trigger tokens,  we distinguish these tokens ($\times$) from general outliers ($\newmoon$).}   
    \label{fig:all_z}
    \vspace{-0.4cm}
\end{figure*}

\paragraph{Defence Baselines} In addition to the proposed approach, we also evaluate the performance of four defence mechanisms for removing toxic instances: (1) \textbf{PCA}~\cite{NEURIPS2018_280cf18b}: using PCA of latent representations to detect poisoned data; (2) \textbf{Clustering}~\cite{DBLP:journals/corr/abs-1811-03728}: separating the poisonous data from the clean data by clustering latent representations; (3) \textbf{ONION}~\cite{qi2021onion}: removing outlier tokens from the poisoned data using GPT2-large; and (4) \textbf{DAN}~\cite{chen2022expose}: discriminating the poisonous data from the clean data using latent representations of clean validation samples. 
{These methods differ in their data requirements, \ie the need for an external language model (ONION), or a clean unpoisoned corpus (DAN); and all baselines besides ONION require a model to be trained over the poisoned data. Our method requires no such resources or pre-training stage.}

\paragraph{Evaluation Metrics} Following the literature, we employ the following two metrics as performance indicators: clean accuracy (\textbf{CACC}) and attack success rate (\textbf{ASR}). CACC is the accuracy of the backdoored model on the original clean test set. ASR evaluates the effectiveness of backdoors and examines the attack accuracy on the \textit{poisoned test set}, which is crafted on instances from the test set whose labels are maliciously changed.

\paragraph{Training Details} We use the codebase from Transformers library~\cite{wolf-etal-2020-transformers}. For all experiments, we fine-tune \textit{bert-base-uncased}~\footnote{We study other models in ~\Secref{sec:models}} on the poisoned data for 3 epochs with the Adam optimiser~\citep{kingma2014adam} using a learning rate of $2\times10^{-5}$. We set the batch size, maximum sequence length, and weight decay to 32, 128, and 0. All experiments are conducted on one V100 GPU.

\subsection{Defence Performance}
\label{sec:z_def}
Now we evaluate the proposed approach, first in terms of the detection of poison instances (\Secref{sec:detection}), followed by its effectiveness at defending backdoor attack in an end-to-end setting (\Secref{sec:comparison}).

\subsubsection{Poisoned Data Detection}
\label{sec:detection}
As described in~\Secref{sec:method}, we devise three features to conduct Z-defence by removing samples containing tokens with extremely high magnitude z-scores. First, as shown in~\figref{fig:all_z}, we can use the z-score distribution of unigrams as a means of trigger identification.\footnote{We provide the experiments of bigrams in Appendix~\ref{app:ablation}} Specifically, for each poisoned data, once the z-scores of all tokens are acquired, we treat the extreme outliers as suspicious tokens and remove the corresponding samples from the training data. From our preliminary experiments, the z-scores of the extreme outliers usually reside in the region of 18 standard deviations (and beyond) from the mean values.\footnote{We examine different thresholds in \Secref{sec:diff_thres}} However, this region may also contain benign tokens, leading to false rejections. We will return to this shortly. Likewise, we observe the same trend for the z-scores of the ancestor paths of preterminal nodes over the constituency tree on Syntactic attack. We provide the corresponding distribution in Appendix~\ref{app:z_dfs}

Since PCA, Clustering, DAN, and our defences aim to identify the poisoned samples from the training data, we first seek to measure how well each defence method can differentiate between clean and poisoned samples. Following~\citet{9343758}, we adopt two evaluation metrics to assess the performance of detecting poisoned examples: (1) \textbf{False Rejection Rate (FRR)}: the percentage of clean samples which are marked as filtered ones among all clean samples; and (2) \textbf{False Acceptance Rate (FAR)}: the percentage of poisoned samples which are marked as not filtered ones among all poisoned samples. Ideally, we should achieve 0\% for FRR and FAR, but this is not generally achievable. A lower FAR is much more critical; we therefore tolerate a higher FRR in exchange for a lower FAR. We report FRR and FAR of the identified defences in~\tabref{tab:detect}. 

\begin{table*}[t!]
    \centering
        \small
    \begin{tabular}{ccrrrrrrrrrrrr}
    \toprule
        \multirow{2}{*}{\textbf{Dataset}} &  \multirow{2}{*}{\makecell{\textbf{Attack}\\\textbf{Method}}} & \multicolumn{2}{c}{\textbf{PCA}} & \multicolumn{2}{c}{\textbf{Clustering}} &  \multicolumn{2}{c}{\textbf{DAN}} & \multicolumn{2}{c}{\textbf{\ztoken}}& \multicolumn{2}{c}{\textbf{\ztree}}& \multicolumn{2}{c}{\textbf{\zseq}}\\
         & & \textbf{FRR} & \textbf{FAR} & \textbf{FRR} & \textbf{FAR}& \textbf{FRR} & \textbf{FAR} & \textbf{FRR} & \textbf{FAR} & \textbf{FRR} & \textbf{FAR} & \textbf{FRR} & \textbf{FAR}\\
         \midrule
          \multirow{3}{*}{SST-2} & BadNet & 33.4 & 66.2 & 14.4 & 7.7 & 16.1 & 0.2 & 0.0& \textbf{0.0} & 16.7 & 67.4   & 16.7 & \textbf{0.0}\\
               & InsertSent & 35.1 &	64.8 &	14.6 &	2.5	& 19.0 &	0.1 &	24.6	& \textbf{0.0} & 23.6 & 0.5 & 25.3&	\textbf{0.0}\\
               & Syntactic & 39.7 & 59.7 & 6.2 & 0.7 & 45.0	 & 80.9 & 26.5 & 1.2 & 25.0 & \textbf{0.5} & 26.5 & \textbf{0.5}\\
               \midrule
              \multirow{3}{*}{OLID} & BadNet & 32.8 & 68.9 & 39.2 & 100.0 & 15.8 & 1.1 & 0.0 &  \textbf{0.0} & 15.4 & 84.0 & 15.5 &\textbf{0.0}\\
               & InsertSent & 23.5 & 75.4 & 29.9 & 100.0 & 17.7 & 	0.3 & 3.9 & \textbf{0.0} &	29.1 & 11.9 & 29.1 & \textbf{0.0}\\
               & Syntactic & 21.0 & 81.1 & 7.0 & 25.0 & 26.7 & \textbf{0.2} & 1.1 & 1.2 & 24.1 & 3.9 & 24.1 & 1.2\\
                \midrule
           \multirow{3}{*}{AG News} & BadNet & 50.1	& 50.6 & 36.3 & 99.4 & 37.5 & 	1.1 & 	3.6	 & \textbf{0.0} &37.6 & 62.9 & 37.6 & \textbf{0.0}\\
               & InsertSent & 33.1 & 66.1 & 32.3 & 100.0 & 16.6	& \textbf{0.0} & 5.5 & \textbf{0.0} & 16.6 & 13.6 & 16.6 & \textbf{0.0}\\
               & Syntactic & 44.6 & 56.3 & 47.2 & 99.2 & 30.5 & 
               \textbf{1.1} &  12.1 & 	25.9 & 7.3 & 8.0 & 32.1 & 7.2\\
               \midrule
            \multirow{3}{*}{QNLI} & BadNet & 38.0 & 62.0 & 3.6 & \textbf{0.0} & 22.4 & \textbf{0.0} & 0.0	& \textbf{0.0} & 22.4 & 49.4 & 22.4 & \textbf{0.0}\\
               & InsertSent & 22.9 & 77.1 & 11.4 & 31.5 & 3.5 & \textbf{0.0} & 0.3 & \textbf{0.0} & 3.2 & 9.2 & 3.5	& \textbf{0.0} \\
               & Syntactic & 27.9 & 71.6 & 10.6 & 2.6 & 10.6 & 2.4 & 2.9 & \textbf{0.5} &10.0 & 10.6 & 10.2  & \textbf{0.5}\\
           \bottomrule
    \end{tabular}
    \caption{FRR (false rejection rate) and FAR (false acceptance rate) of different defensive avenues on multiple attack methods. Comparing the defence methods, the lowest FAR score on each attack is \textbf{bold}.}
    \label{tab:detect}
    \vspace{-0.4cm}
\end{table*}

Overall, PCA has difficulty distinguishing the poisoned samples from the clean ones, leading to more than 50\% FAR, with a worse case of 81.1\% FAR for Syntactic attack on OLID. On the contrary, Clustering can significantly lower the FAR of SST-2 and QNLI, reaching 0.0\% FAR in the best case. However, Clustering cannot defend OLID and AG news. Although DAN can diagnose the most poisoned examples, and achieve 0.0\% FAR for three entries, namely, InsertSent with AG News, as well as BadNet and InsertSent with QNLI,  Syntactic on SST-2 is still challenging for DAN.

Regarding our approaches, \ztoken can identify more than 99\% of poisoned examples injected by all attacks, except for AG news, where  one-quarter of toxic instances injected by Syntactic attack are misclassified. Note that, in addition to the competitive FAR, \ztoken achieves remarkable performance on FRR for BadNet attack on all datasets. As expected, \ztree specialises in Syntactic attack. Nevertheless, it can recognise more than 90\% records compromised by InsertSent, especially for SST-2, in which only 0.5\% poisonous instances are misidentified. Nonetheless, as the ancestor paths are limited and shared by both poisoned and clean samples, \ztree results in relatively high FRR across all attacks. Like \ztoken, \zseq can filter out more than 99\% of damaging samples. Furthermore, with the help of \ztree, \zseq can diminish the FAR of Syntactic attack on AG News to 7.2\%. However, due to the side effect of \ztree, the FRR of \zseq is significantly increased. Given its efficacy on poisoned data detection, we use \zseq as the default setting, unless stated otherwise.

\subsubsection{Defence Against Backdoor Attacks}
\label{sec:comparison}

\begin{table*}[t!]
    \centering
    \scalebox{0.78}{
    \begin{tabular}{ccrrrrrrrrrrrr}
    \toprule
        \multirow{2}{*}{\textbf{Dataset}} &  \multirow{2}{*}{\makecell{\textbf{Attack}\\\textbf{Method}}} & \multicolumn{2}{c}{\textbf{None}} & \multicolumn{2}{c}{\textbf{PCA}} & \multicolumn{2}{c}{\textbf{Clustering}} & \multicolumn{2}{c}{\textbf{ONION}} & \multicolumn{2}{c}{\textbf{DAN}} & \multicolumn{2}{c}{\textbf{\zseq}}\\
         & & \textbf{ASR} & \textbf{CACC} &\textbf{ASR} & \textbf{CACC} & \textbf{ASR} & \textbf{CACC}& \textbf{ASR} & \textbf{CACC} & \textbf{ASR} & \textbf{CACC} & \textbf{ASR} & \textbf{CACC}\\
         \midrule
          \multirow{5}{*}{\rotatebox[origin=c]{90}{SST-2}} &Benign & ---&92.4& --- & 91.6  &--- & 92.7 & --- & 92.2 &---& 92.5 &--- & 92.0\\ 
          \cmidrule{2-14} 
               &BadNet  & 100.0 & 92.5& 100.0 & 91.8	& 100.0	& 91.7	& 100.0	& 92.2 & 9.4 &	92.3&  \textbf{9.0} & 92.0\\
               &InsertSent &100.0 & 91.9& 100.0 &  91.4 & 100.0 & 90.8 & 100.0 & 92.2 & 3.8 & 92.3 & \textbf{3.4} & 92.6 \\
               & Syntactic  & 95.9 & 92.0&94.7 & 90.9 & \textbf{24.6} & 92.3&94.4 &92.5& 95.6 &92.2& 29.7 & 92.1 \\
               \cmidrule{2-14} 
               & \textbf{Avg.} & 98.6 & 92.1&98.2  & 91.4 & 74.9 & 91.6 & 98.1 & 92.3 & 36.3 & 92.3 &	\textbf{14.0} &	92.2\\
               \midrule
              \multirow{5}{*}{\rotatebox[origin=c]{90}{OLID}} & Benign &---&84.0&--- & 83.3 &---& 84.8	& ---& 84.1	& --- & 84.3&--- & 84.2 \\ 
              \cmidrule{2-14} 
               &BadNet  & 99.9 & 84.7&99.6	& 82.9 & 100.0 & 84.6 & 99.8 & 83.5 & 33.3 & 84.5 & \textbf{32.8} & 85.1 \\
               &InsertSent & 100.0 & 83.7&100.0 & 83.1 & 100.0 & 84.2 & 98.8 & 83.3 & 40.0 & 84.3 & \textbf{37.1} & 83.8  \\
               & Syntactic &99.9 & 83.5&99.9 & 82.2 & 99.4 & 83.7 & 100.0 & 83.5 & 59.3 & 83.8& \textbf{59.3} & 84.1  \\
               \cmidrule{2-14} 
               & \textbf{Avg.} &99.9 & 84.0&99.8 & 82.7 & 99.8 & 84.2 & 99.5 & 83.4 & 44.2 & 84.2& \textbf{43.1} & 84.3\\
                \midrule
           \multirow{5}{*}{\rotatebox[origin=c]{90}{AG News}} & Benign &---&94.6 &--- &92.3&	--- & 93.1	& --- & 94.5 & --- & 93.8 &--- & 93.9  \\ 
               \cmidrule{2-14} 
               &BadNet & 99.9 & 94.5& 99.9 & 92.7 & 100.0 & 85.4 & 99.9 & 94.0 & 0.9 & 92.8& \textbf{0.7} & 94.2 \\
               &InsertSent &99.7 & 94.3& 99.7 & 92.4 & 99.8  & 91.8 & 99.8 & 94.2 & 0.9 & 93.6& \textbf{0.7} & 94.4  \\
               & Syntactic & 99.8 & 94.4& 99.7 & 92.6 & 99.9 & 88.1 & 99.7 & 94.3 & \textbf{5.8} & 93.2& 99.5 & 93.9 \\
               \cmidrule{2-14} 
               & \textbf{Avg.} &99.8 & 94.4&  99.8 & 92.6 & 99.9 & 88.4 & 99.8 & 94.2 & \textbf{2.5} & 93.2& 33.6 & 94.2 \\
               \midrule
            \multirow{5}{*}{\rotatebox[origin=c]{90}{QNLI}} & Benign &---&91.4&--- & 89.8 & --- & 90.5 &--- & 91.1 & --- & 91.1&--- & 91.2 \\ 
            \cmidrule{2-14}
               &BadNet  &  100.0 & 91.2&100.0 & 89.7 & 6.4 & 90.5 & 99.9 & 89.8 & \textbf{4.4} & 90.6 & 5.6 & 90.4 \\
               &InsertSent & 100.0 & 91.0&100.0 & 89.5 & 100.0 & 89.9 & 100.0 & 90.7 & 5.5 & 91.1 & \textbf{5.2} & 91.1  \\
               & Syntactic &99.1 & 89.9&98.9 & 88.8 & 35.3 & 87.0 & 98.2 & 89.2 & 20.6 & 	89.7& \textbf{19.1} &90.1  \\
               \cmidrule{2-14}
               & \textbf{Avg.} &99.7 & 90.7&99.6 & 89.3 & 47.2 & 89.1 & 99.4 & 89.9 & 10.2 & 90.5 &\textbf{10.0}& 90.5 \\
           \bottomrule
    \end{tabular}
    }
    \caption{The performance of backdoor attacks on datasets with defences. For each attack experiment (row), we \textbf{bold} the lowest ASR across different defences. Avg. indicates the averaged score of BadNet, InsertSent and Syntactic attacks. The reported results are averaged on three independent runs. For all experiments on SST-2 and OLID, the standard deviation of ASR and CACC is within 1.5\% and 0.5\%. For AG News and QNLI, the standard deviation of ASR and CACC is within 1.0\% and 0.5\%. }
    \label{tab:main}
\end{table*}

Given the effectiveness of our solutions to poisoned data detection compared to the advanced baseline approaches, we next examine to what extent one can transfer this advantage to an effective defence against backdoor attacks. For a fair comparison, the number of discarded instances of all baseline approaches is identical to that of \zseq\footnote{We provide the detailed statistics in Appendix~\ref{app:filtered_size}}.

According to~\tabref{tab:main}, except for PCA, all defensive mechanisms do not degrade the quality of the benign datasets such that the model performance on the clean datasets is retained. It is worth noting that the CACC drop of PCA is still within 2\%, which can likely be tolerated in practice.

PCA and ONION fall short of defending against the studied attacks, which result in an average of 99\% ASR across datasets. Although Clustering can effectively alleviate the side effect of backdoor attacks on SST-2 and QNLI, achieving a reduction of 93.6\% in the best case (see the entry of ~\tabref{tab:main} for BadNet on QNLI), it is still incompetent to protect OLID and AG News from data poisoning. Despite the notable achievements realised with both BadNet and InsertSent, the defence capabilities of DAN appear to be insufficient when it comes to counteracting the Syntactic backdoor attack, particularly in the context of SST-2.

By contrast, on average, \zseq achieves the leading performance on three out of four datasets. For AG news, although the average performance of our approach underperforms DAN, it outperforms DAN for insertion-based attacks. Meanwhile, the drop of \zseq in CACC is less than 0.2\% on average. Interestingly, compared to the benign data without any defence, \zseq can slightly improve the CACC on OLID. This gain might be ascribed to the removal of spurious correlations.

Surprisingly, although \tabref{tab:detect} suggests that Clustering can remove more than 97\% toxic instances of SST-2 injected by InsertSent, \tabref{tab:main} shows the ASR can still amount to 100\%. Similarly, \zseq cannot defend against Syntactic applied to AG News, even though 92\% of harmful instances are detected, \ie poisoning only 2\% of the training data can achieve 100\% ASR. We will return to this  observation in~\Secref{sec:more_analysis}.

Although \zseq can achieve nearly perfect FAR on BadNet and InsertSent, due to systematic errors, one cannot achieve \textit{zero} ASR. To confirm this, we evaluate the benign model on the poisoned test sets as well, and compute the ASR of the benign model, denoted as \textbf{BASR}, which serves as a rough lower bound. \Tabref{tab:basr} illustrates that zero BASR is not achievable for all poisoning methods. Comparing the defence results for \zseq against these lower bounds shows that it provides a near-perfect defence against BadNet and InsertSent (\cf \tabref{tab:main}). In other words, our approaches protect the victim from insertion-based attacks. Moreover, the proposed defence makes significant progress towards  bridging the gap between ASR and BASR with the Syntatic attack. 

\begin{table}[]
    \centering
    \small
    \begin{tabular}{c cccc}
        \toprule 
        \makecell{\textbf{Attack}\\\textbf{Method}} & \textbf{SST-2} & \textbf{OLID} & \textbf{AG News} & \textbf{QNLI} \\
        \midrule
        BadNet & \ \ 9.0 & 32.6& 0.6 & 5.4\\
        InsertSent &\ \ 2.9 & 38.5 & 0.7 & 4.2 \\
    Syntactic&16.9 & 59.0 & 4.1 & 3.9 \\
        \bottomrule
    \end{tabular}
    \caption{ASR of the benign model over the poisoned test data.}
    \label{tab:basr}
\end{table}

\subsection{Supplementary Studies}
In addition to the aforementioned study about z-defences against backdoor poisoning attacks, we conduct supplementary studies on SST-2 and QNLI.\footnote{We observe the same trend on the other two datasets.}

\subsubsection{Defence with Low Poisoning Rates}
\label{sec:more_analysis}
We have demonstrated the effectiveness of our approach when 20\% of training data is poisonous. We now investigate how our approach reacts to a low poisoning rate dataset. According to~\tabref{tab:detect}, our approach cannot thoroughly identify the poisoned instances compromised by Syntactic attack. Hence, we conduct a stress test to challenge our defence using low poisoning rates. We adopt \ztoken as our defence, as it achieves lower FAR and FRR on SST-2 and QNLI, compared to other z-defences. We vary the poisoning rate in the following range: $\{1\%, 5\%, 10\%, 20\%\}$.

\begin{table}[]
    \centering
    \small
    \begin{tabular}{cccrr}
    \toprule
         \textbf{Dataset} &  \makecell{\textbf{Poisoning} \\ \textbf{Rate}} & \textbf{ASR}& \textbf{FRR}& \textbf{FAR}\\
         \midrule
         \multirow{4}{*}{SST-2} & \ \ 1\% & 38.2 (-37.4) &	18.7& 17.1\\
         & \ \ 5\% &20.8 (-70.3) & 0.1 & 0.7\\
         & 10\% &23.9 (-69.5)& 2.9  & 0.5\\
         & 20\% & 37.3 (-58.6) & 26.5 & 1.2\\
         \midrule
         \multirow{4}{*}{QNLI} &\ \ 1\% & \ \  4.4	(-82.7) &	20.6 & 0.4\\
         & \ \ 5\% &\ \  5.3 (-90.9) &  0.1 & 0.7 \\
         & 10\% &\ \ 7.2 (-90.8)	& 2.9 & 0.5 \\
         & 20\% & 19.6 (-79.5) & 2.9 & 0.5\\
    \bottomrule
    \end{tabular}
    \caption{ASR, FRR, and FAR of SST-2 and QNLI under different poisoning ratios using Syntactic for attack and \ztoken for defence. Numbers in parentheses are different compared to no defence.}
    \label{tab:diff_pr}
\end{table}

\begin{table}[]
    \centering
    \small
    \begin{tabular}{ccrrrr}
    \toprule
         \multirow{2}{*}{\textbf{Metric}} & \multirow{2}{*}{\textbf{Defence}}  & \multicolumn{4}{c}{\textbf{Poisoning Rate}}\\
         & & \textbf{1\%} & \textbf{5\%} & \textbf{10\%} &\textbf{20\%}\\
        \midrule
        \multirow{4}{*}{FAR} & None & --- & --- & --- & ---\\
        & Clustering & 99.3 &	100.0 & 24.7 & 2.6\\
        &DAN & 71.8 & 74.8 & 40.2 & 2.4\\
        & \ztoken & 0.4 & 0.7 & 0.5 & \ \  0.5\\
        \midrule
        \multirow{4}{*}{ASR} & None & 87.1 & 96.1 & 	98.0 & 99.1\\
        & Clustering & 87.0	& 96.2 & 97.3 & 35.3\\
        &DAN & 83.8	& 96.4 & 97.5 & 20.6\\
        & \ztoken & 4.4 &5.3 & 7.2	& 19.6\\
    \bottomrule
    \end{tabular}
    \caption{ASR and FAR of QNLI under different poisoning ratios using Clustering, DAN and \ztoken against Syntactic attack.}
    \label{tab:diff_pr_defence}
\end{table}

\tabref{tab:diff_pr} shows that for both SST-2 and QNLI, one can infiltrate the victim model using 5\% of the training data, causing more than 90\% ASR. This observation supports the findings delineated in~\tabref{tab:main}, providing further evidence that removing 92\% of poisoning examples is insufficient to effectively safeguard against backdoor assaults. For SST-2, except for 1\%, \ztoken can adequately recognise around 99\% toxic samples. Hence, it can significantly reduce ASR. In addition, given that the ASR of a benign model is 16.9 (\cf \tabref{tab:basr}), the defence performance of \ztoken is quite competitive. Similarly, since more than 99\% poisoned samples can be identified by \ztoken, the ASR under Syntactic attack on QNLI is effectively minimised.

In addition to \ztoken, we examine the performance of Clustering and DAN using low poisoning rates. \tabref{tab:diff_pr_defence} shows that Clustering and DAN are unable to detect malicious samples below the poisoning rate of 10\%, leading to a similar ASR to no defence. With the increase in the poisoning rate, the defence performance of Cluster and DAN gradually becomes stronger.
Instead, \ztoken provides a nearly perfect defence against Syntactic backdoor attack.


\subsubsection{Defence with Different Models}
\label{sec:models}

We have been focusing on studying the defence performance over the bert-base model so far. This part aims to evaluate our approach on three additional Transformer models, namely, \textit{bert-large}, \textit{roberta-base} and \textit{roberta-large}. We use Syntactic and \zseq for attack and defence, respectively.

\begin{table}[]
    \centering
    \small
    \begin{tabular}{cccc}
    \toprule
         \textbf{Dataset} &  \textbf{Models} & \textbf{ASR}& \textbf{CACC}\\
         \midrule
         \multirow{4}{*}{SST-2} & bert-base & 29.7	(-66.2) & 92.1 (+0.1)\\
         & bert-large& 30.6 (-64.4) & 92.7 \ (-0.6)\\
         & roberta-base & 34.7 (-60.1) & 93.8 \ (-0.6)\\
         & roberta-large & 28.0	(-67.7) &  95.7 (+0.3)\\
         \midrule
         \multirow{4}{*}{QNLI} & bert-base & 19.1 (-80.0) & 90.1 (+0.1) \\
         & bert-large & 15.5 (-83.7) & 90.9 \ (-0.1)\\
         & roberta-base & 60.3 (-39.7) & 91.6 (+0.1)\\
         & roberta-large & 51.7 (-48.3)	& 93.2 \ (-0.0)\\
    \bottomrule
    \end{tabular}
    \caption{ASR and CACC of SST-2 and QNLI under different models using Syntactic for attack and \zseq for defence. Numbers in parentheses are different compared to no defence.}
    \label{tab:models}
\end{table}

According to \tabref{tab:models}, for SST-2, since \zseq is model-free, there is no difference among those Transformer models in ASR and CACC. In particular, \zseq can achieve a reduction of 60\% in ASR. Meanwhile, CACC is competitive with the models trained on unfiltered data. Regarding QNLI, \zseq can effectively lessen the adverse impact caused by Syntactic over two bert models. Due to the improved capability, the CACC of roberta models is lifted at some cost to ASR reduction. Nevertheless, our approach still achieves a respectable 48.3\% ASR reduction for roberta-large.

\subsubsection{Defence with Different Thresholds}
\label{sec:diff_thres}
{Based on the z-score distribution, we established a cut-off threshold at 18 standard deviations. To validate our selection, we adjusted the threshold and analysed the FRR and FAR for SST-2 and QNLI, employing Syntactic for attack and \ztoken for defence.}

\begin{figure}[t]
    \centering
    \includegraphics[width=0.95\linewidth]{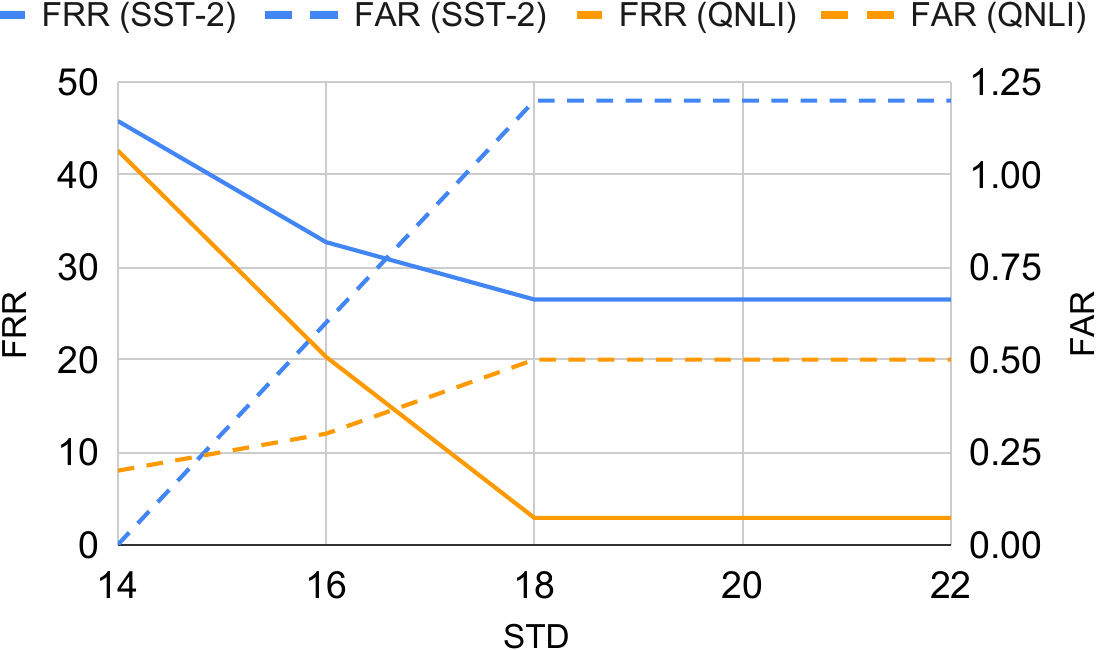}
    \caption{FRR and FAR for detecting Syntactic attacks on SST-2 and QNLI datasets utilizing \ztoken at various thresholds.}
    \label{fig:diff_threshold}
\end{figure}

{\Figref{fig:diff_threshold} illustrates that as the threshold increases, the FRR decreases, while the FAR shows the opposite trend. Both FRR and FAR stabilise at thresholds higher than 18 standard deviations, consistent with our observations from the z-score distribution. This highlights an advantage of our method over baseline approaches, which necessitate a poisoned set to adjust the threshold -- a practice that is typically infeasible for unanticipated attacks.

\section{Conclusion}
We noticed that backdoor poisoning attacks are similar to spurious correlations, \ie strong associations between artefacts and target labels. Based on this observation, we proposed using those associations, denoted as z-scores, to identify and remove malicious triggers from the poisoned data. Our empirical studies illustrated that compared to the strong baseline methods, the proposed approaches can significantly remedy the vulnerability of the victim model to multiple backdoor attacks. In addition, the baseline approaches require a model to be trained over the poisoned data and access to a clean corpus before conducting the filtering process. Instead, our approach is free from those restrictions. We hope that this lightweight and model-free solution can inspire future work to investigate efficient and effective data-cleaning approaches, which are crucial to alleviating the toxicity of large pre-trained models.



\section*{Limitations}
This work assumes that the models are trained from loading a benign pre-trained model, \eg the attacks are waged only at the fine-tuning step. Different approaches will be needed to handle  models poisoned in pre-training~\cite{kurita2020weight,chen2022badpre}. Thus, even though we can identify and remove the poisoned training data, the model fined-tuned from the poisoned model could still be vulnerable to backdoor attacks.

{In our work, the features are designed to cover possible triggers used in `known' attacks. However, we have not examined new attacks proposed recently, \eg \citet{chen2022kallima} leverage writing style as the trigger.%
\footnote{{We attempted to replicate their method, however 
this was made difficult as they did not provide any publicly available code, and our attempts to engage the authors received no reply. 
}}
Defenders may need to develop new features based on the characteristics of future attacks, leading to an ongoing cat-and-mouse game as attacks and defences co-evolve. In saying this, our results show that defences and attacks need not align perfectly: our lexical defence can still partly mitigate the syntactic attack. Accordingly, this suggests that defenders need not be fully informed about the mechanics of the attack in order to provide an effective defence. Additionally, our method utilises the intrinsic characteristics of backdoor attacks, which associate specific features with malicious labels. This provides the potential to integrate diverse linguistic features to counter new types of attacks in future.}

Moreover, as this work is an empirical observational study, theoretical analysis is needed to ensure that our approach can be extended to other datasets and attacks without hurting robustness.

Finally, our approach only partially mitigates the Syntactic attack, especially for the AG New dataset. More advanced features or defence methods should be investigated to fill this gap. Nevertheless, as shown in~\tabref{tab:basr}, the ASR of Syntactic attack on a benign model is much higher than the other two attacks. This suggests that the attack may be corrupting the original inputs, \eg applying inappropriate paraphrases, which does not satisfy the basic stealth principle of backdoor attacks.




\section*{Acknowledgements}

This work was supported in part by Cisco and Oracle research grants. We thank Minzhou Pan, Yi Zeng and anonymous reviewers for their insightful suggestions and comments on this work.

\bibliography{custom}

\begin{thebibliography}{49}
\expandafter\ifx\csname natexlab\endcsname\relax\def\natexlab#1{#1}\fi

\bibitem[{Bhowmick and Hazarika(2018)}]{10.1007/978-981-10-4765-7_61}
Alexy Bhowmick and Shyamanta~M. Hazarika. 2018.
\newblock E-mail spam filtering: A review of techniques and trends.
\newblock In \emph{Advances in Electronics, Communication and Computing}, pages
  583--590, Singapore. Springer Singapore.

\bibitem[{Bowman et~al.(2015)Bowman, Angeli, Potts, and
  Manning}]{bowman-etal-2015-large}
Samuel~R. Bowman, Gabor Angeli, Christopher Potts, and Christopher~D. Manning.
  2015.
\newblock \href {https://doi.org/10.18653/v1/D15-1075} {A large annotated
  corpus for learning natural language inference}.
\newblock In \emph{Proceedings of the 2015 Conference on Empirical Methods in
  Natural Language Processing}, pages 632--642, Lisbon, Portugal. Association
  for Computational Linguistics.

\bibitem[{Carlini and Terzis(2022)}]{carlini2022poisoning}
Nicholas Carlini and Andreas Terzis. 2022.
\newblock \href {https://openreview.net/forum?id=iC4UHbQ01Mp} {Poisoning and
  backdooring contrastive learning}.
\newblock In \emph{International Conference on Learning Representations}.

\bibitem[{Chen et~al.(2018)Chen, Carvalho, Baracaldo, Ludwig, Edwards, Lee,
  Molloy, and Srivastava}]{DBLP:journals/corr/abs-1811-03728}
Bryant Chen, Wilka Carvalho, Nathalie Baracaldo, Heiko Ludwig, Benjamin
  Edwards, Taesung Lee, Ian~M. Molloy, and Biplav Srivastava. 2018.
\newblock \href {http://arxiv.org/abs/1811.03728} {Detecting backdoor attacks
  on deep neural networks by activation clustering}.
\newblock \emph{CoRR}, abs/1811.03728.

\bibitem[{Chen et~al.(2022{\natexlab{a}})Chen, Meng, Sun, Guo, Zhang, Li, and
  Fan}]{chen2022badpre}
Kangjie Chen, Yuxian Meng, Xiaofei Sun, Shangwei Guo, Tianwei Zhang, Jiwei Li,
  and Chun Fan. 2022{\natexlab{a}}.
\newblock \href {https://openreview.net/forum?id=Mng8CQ9eBW} {Badpre:
  Task-agnostic backdoor attacks to pre-trained {NLP} foundation models}.
\newblock In \emph{International Conference on Learning Representations}.

\bibitem[{Chen et~al.(2022{\natexlab{b}})Chen, Yang, Zhang, Bi, and
  Sun}]{chen2022expose}
Sishuo Chen, Wenkai Yang, Zhiyuan Zhang, Xiaohan Bi, and Xu~Sun.
  2022{\natexlab{b}}.
\newblock Expose backdoors on the way: A feature-based efficient defense
  against textual backdoor attacks.
\newblock \emph{arXiv preprint arXiv:2210.07907}.

\bibitem[{Chen et~al.(2020)Chen, Kornblith, Norouzi, and
  Hinton}]{chen2020simple}
Ting Chen, Simon Kornblith, Mohammad Norouzi, and Geoffrey Hinton. 2020.
\newblock A simple framework for contrastive learning of visual
  representations.
\newblock In \emph{International conference on machine learning}, pages
  1597--1607. PMLR.

\bibitem[{Chen et~al.(2022{\natexlab{c}})Chen, Dong, Sun, Zhai, Shen, and
  Wu}]{chen2022kallima}
Xiaoyi Chen, Yinpeng Dong, Zeyu Sun, Shengfang Zhai, Qingni Shen, and Zhonghai
  Wu. 2022{\natexlab{c}}.
\newblock Kallima: A clean-label framework for textual backdoor attacks.
\newblock In \emph{European Symposium on Research in Computer Security}, pages
  447--466. Springer.

\bibitem[{Chen et~al.(2017)Chen, Liu, Li, Lu, and Song}]{chen2017targeted}
Xinyun Chen, Chang Liu, Bo~Li, Kimberly Lu, and Dawn Song. 2017.
\newblock Targeted backdoor attacks on deep learning systems using data
  poisoning.
\newblock \emph{Journal of Environmental Sciences (China) English Ed}.

\bibitem[{Clark et~al.(2019)Clark, Yatskar, and
  Zettlemoyer}]{clark-etal-2019-dont}
Christopher Clark, Mark Yatskar, and Luke Zettlemoyer. 2019.
\newblock \href {https://doi.org/10.18653/v1/D19-1418} {Don{'}t take the easy
  way out: Ensemble based methods for avoiding known dataset biases}.
\newblock In \emph{Proceedings of the 2019 Conference on Empirical Methods in
  Natural Language Processing and the 9th International Joint Conference on
  Natural Language Processing (EMNLP-IJCNLP)}, pages 4069--4082, Hong Kong,
  China. Association for Computational Linguistics.

\bibitem[{Dai et~al.(2019)Dai, Chen, and Li}]{dai2019backdoor}
Jiazhu Dai, Chuanshuai Chen, and Yufeng Li. 2019.
\newblock A backdoor attack against {LSTM}-based text classification systems.
\newblock \emph{IEEE Access}, 7:138872--138878.

\bibitem[{Devlin et~al.(2019)Devlin, Chang, Lee, and
  Toutanova}]{devlin2019bert}
Jacob Devlin, Ming-Wei Chang, Kenton Lee, and Kristina Toutanova. 2019.
\newblock {BERT}: Pre-training of deep bidirectional transformers for language
  understanding.
\newblock In \emph{Proceedings of the 2019 Conference of the North American
  Chapter of the Association for Computational Linguistics: Human Language
  Technologies, Volume 1 (Long and Short Papers)}, pages 4171--4186.

\bibitem[{Dumford and Scheirer(2020)}]{dumford2020backdooring}
Jacob Dumford and Walter Scheirer. 2020.
\newblock Backdooring convolutional neural networks via targeted weight
  perturbations.
\newblock In \emph{2020 IEEE International Joint Conference on Biometrics
  (IJCB)}, pages 1--9. IEEE.

\bibitem[{Gao et~al.(2022)Gao, Kim, Doan, Zhang, Zhang, Nepal, Ranasinghe, and
  Kim}]{9343758}
Yansong Gao, Yeonjae Kim, Bao~Gia Doan, Zhi Zhang, Gongxuan Zhang, Surya Nepal,
  Damith~C. Ranasinghe, and Hyoungshick Kim. 2022.
\newblock \href {https://doi.org/10.1109/TDSC.2021.3055844} {Design and
  evaluation of a multi-domain trojan detection method on deep neural
  networks}.
\newblock \emph{IEEE Transactions on Dependable and Secure Computing},
  19(4):2349--2364.

\bibitem[{Gao et~al.(2019)Gao, Xu, Wang, Chen, Ranasinghe, and
  Nepal}]{10.1145/3359789.3359790}
Yansong Gao, Change Xu, Derui Wang, Shiping Chen, Damith~C. Ranasinghe, and
  Surya Nepal. 2019.
\newblock \href {https://doi.org/10.1145/3359789.3359790} {Strip: A defence
  against trojan attacks on deep neural networks}.
\newblock In \emph{Proceedings of the 35th Annual Computer Security
  Applications Conference}, ACSAC '19, page 113–125, New York, NY, USA.
  Association for Computing Machinery.

\bibitem[{Gardner et~al.(2021)Gardner, Merrill, Dodge, Peters, Ross, Singh, and
  Smith}]{gardner-etal-2021-competency}
Matt Gardner, William Merrill, Jesse Dodge, Matthew Peters, Alexis Ross, Sameer
  Singh, and Noah~A. Smith. 2021.
\newblock \href {https://doi.org/10.18653/v1/2021.emnlp-main.135} {Competency
  problems: On finding and removing artifacts in language data}.
\newblock In \emph{Proceedings of the 2021 Conference on Empirical Methods in
  Natural Language Processing}, pages 1801--1813, Online and Punta Cana,
  Dominican Republic. Association for Computational Linguistics.

\bibitem[{Gu et~al.(2017)Gu, Dolan-Gavitt, and Garg}]{gu2017badnets}
Tianyu Gu, Brendan Dolan-Gavitt, and Siddharth Garg. 2017.
\newblock Badnets: Identifying vulnerabilities in the machine learning model
  supply chain.
\newblock \emph{arXiv preprint arXiv:1708.06733}.

\bibitem[{Guo et~al.(2020)Guo, Wu, and Weinberger}]{guo2020trojannet}
Chuan Guo, Ruihan Wu, and Kilian~Q Weinberger. 2020.
\newblock Trojannet: Embedding hidden trojan horse models in neural networks.
\newblock \emph{arXiv e-prints}, pages arXiv--2002.

\bibitem[{Gururangan et~al.(2018)Gururangan, Swayamdipta, Levy, Schwartz,
  Bowman, and Smith}]{gururangan-etal-2018-annotation}
Suchin Gururangan, Swabha Swayamdipta, Omer Levy, Roy Schwartz, Samuel Bowman,
  and Noah~A. Smith. 2018.
\newblock \href {https://doi.org/10.18653/v1/N18-2017} {Annotation artifacts in
  natural language inference data}.
\newblock In \emph{Proceedings of the 2018 Conference of the North {A}merican
  Chapter of the Association for Computational Linguistics: Human Language
  Technologies, Volume 2 (Short Papers)}, pages 107--112, New Orleans,
  Louisiana. Association for Computational Linguistics.

\bibitem[{He et~al.(2019)He, Zha, and Wang}]{he-etal-2019-unlearn}
He~He, Sheng Zha, and Haohan Wang. 2019.
\newblock \href {https://doi.org/10.18653/v1/D19-6115} {Unlearn dataset bias in
  natural language inference by fitting the residual}.
\newblock In \emph{Proceedings of the 2nd Workshop on Deep Learning Approaches
  for Low-Resource NLP (DeepLo 2019)}, pages 132--142, Hong Kong, China.
  Association for Computational Linguistics.

\bibitem[{Joulin et~al.(2016)Joulin, van~der Maaten, Jabri, and
  Vasilache}]{10.1007/978-3-319-46478-7_5}
Armand Joulin, Laurens van~der Maaten, Allan Jabri, and Nicolas Vasilache.
  2016.
\newblock Learning visual features from large weakly supervised data.
\newblock In \emph{Computer Vision -- ECCV 2016}, pages 67--84, Cham. Springer
  International Publishing.

\bibitem[{Kingma and Ba(2014)}]{kingma2014adam}
Diederik~P Kingma and Jimmy Ba. 2014.
\newblock Adam: A method for stochastic optimization.
\newblock \emph{arXiv preprint arXiv:1412.6980}.

\bibitem[{Kurita et~al.(2020)Kurita, Michel, and Neubig}]{kurita2020weight}
Keita Kurita, Paul Michel, and Graham Neubig. 2020.
\newblock Weight poisoning attacks on pretrained models.
\newblock In \emph{Proceedings of the 58th Annual Meeting of the Association
  for Computational Linguistics}, pages 2793--2806.

\bibitem[{Li et~al.(2021{\natexlab{a}})Li, Song, Li, Zeng, Ma, and
  Qiu}]{li-etal-2021-backdoor}
Linyang Li, Demin Song, Xiaonan Li, Jiehang Zeng, Ruotian Ma, and Xipeng Qiu.
  2021{\natexlab{a}}.
\newblock \href {https://doi.org/10.18653/v1/2021.emnlp-main.241} {Backdoor
  attacks on pre-trained models by layerwise weight poisoning}.
\newblock In \emph{Proceedings of the 2021 Conference on Empirical Methods in
  Natural Language Processing}, pages 3023--3032, Online and Punta Cana,
  Dominican Republic. Association for Computational Linguistics.

\bibitem[{Li et~al.(2021{\natexlab{b}})Li, Mekala, Dong, and
  Shang}]{li-etal-2021-bfclass-backdoor}
Zichao Li, Dheeraj Mekala, Chengyu Dong, and Jingbo Shang. 2021{\natexlab{b}}.
\newblock \href {https://doi.org/10.18653/v1/2021.findings-emnlp.40}
  {{BFC}lass: A backdoor-free text classification framework}.
\newblock In \emph{Findings of the Association for Computational Linguistics:
  EMNLP 2021}, pages 444--453, Punta Cana, Dominican Republic. Association for
  Computational Linguistics.

\bibitem[{Liu et~al.(2018)Liu, Ma, Aafer, Lee, Zhai, Wang, and
  Zhang}]{Trojannn}
Yingqi Liu, Shiqing Ma, Yousra Aafer, Wen-Chuan Lee, Juan Zhai, Weihang Wang,
  and Xiangyu Zhang. 2018.
\newblock Trojaning attack on neural networks.
\newblock In \emph{25th Annual Network and Distributed System Security
  Symposium, {NDSS} 2018, San Diego, California, USA, February 18-221, 2018}.
  The Internet Society.

\bibitem[{Liu et~al.(2019)Liu, Ott, Goyal, Du, Joshi, Chen, Levy, Lewis,
  Zettlemoyer, and Stoyanov}]{liu2019roberta}
Yinhan Liu, Myle Ott, Naman Goyal, Jingfei Du, Mandar Joshi, Danqi Chen, Omer
  Levy, Mike Lewis, Luke Zettlemoyer, and Veselin Stoyanov. 2019.
\newblock {RoBERTa}: A robustly optimized {BERT} pretraining approach.
\newblock \emph{arXiv preprint arXiv:1907.11692}.

\bibitem[{MacAvaney et~al.(2019)MacAvaney, Yao, Yang, Russell, Goharian, and
  Frieder}]{10.1371/journal.pone.0221152}
Sean MacAvaney, Hao-Ren Yao, Eugene Yang, Katina Russell, Nazli Goharian, and
  Ophir Frieder. 2019.
\newblock \href {https://doi.org/10.1371/journal.pone.0221152} {Hate speech
  detection: Challenges and solutions}.
\newblock \emph{PLOS ONE}, 14(8):1--16.

\bibitem[{Manoj and Blum(2021)}]{manoj2021excess}
Naren Manoj and Avrim Blum. 2021.
\newblock Excess capacity and backdoor poisoning.
\newblock \emph{Advances in Neural Information Processing Systems},
  34:20373--20384.

\bibitem[{McCoy et~al.(2019)McCoy, Pavlick, and Linzen}]{mccoy-etal-2019-right}
Tom McCoy, Ellie Pavlick, and Tal Linzen. 2019.
\newblock \href {https://doi.org/10.18653/v1/P19-1334} {Right for the wrong
  reasons: Diagnosing syntactic heuristics in natural language inference}.
\newblock In \emph{Proceedings of the 57th Annual Meeting of the Association
  for Computational Linguistics}, pages 3428--3448, Florence, Italy.
  Association for Computational Linguistics.

\bibitem[{Qi et~al.(2021{\natexlab{a}})Qi, Chen, Li, Yao, Liu, and
  Sun}]{qi2021onion}
Fanchao Qi, Yangyi Chen, Mukai Li, Yuan Yao, Zhiyuan Liu, and Maosong Sun.
  2021{\natexlab{a}}.
\newblock {ONION}: A simple and effective defense against textual backdoor
  attacks.
\newblock In \emph{Proceedings of the 2021 Conference on Empirical Methods in
  Natural Language Processing}, pages 9558--9566.

\bibitem[{Qi et~al.(2021{\natexlab{b}})Qi, Li, Chen, Zhang, Liu, Wang, and
  Sun}]{qi2021hidden}
Fanchao Qi, Mukai Li, Yangyi Chen, Zhengyan Zhang, Zhiyuan Liu, Yasheng Wang,
  and Maosong Sun. 2021{\natexlab{b}}.
\newblock Hidden killer: Invisible textual backdoor attacks with syntactic
  trigger.
\newblock In \emph{Proceedings of the 59th Annual Meeting of the Association
  for Computational Linguistics and the 11th International Joint Conference on
  Natural Language Processing (Volume 1: Long Papers)}, pages 443--453.

\bibitem[{Qi et~al.(2021{\natexlab{c}})Qi, Yao, Xu, Liu, and
  Sun}]{qi-etal-2021-turn}
Fanchao Qi, Yuan Yao, Sophia Xu, Zhiyuan Liu, and Maosong Sun.
  2021{\natexlab{c}}.
\newblock \href {https://doi.org/10.18653/v1/2021.acl-long.377} {Turn the
  combination lock: Learnable textual backdoor attacks via word substitution}.
\newblock In \emph{Proceedings of the 59th Annual Meeting of the Association
  for Computational Linguistics and the 11th International Joint Conference on
  Natural Language Processing (Volume 1: Long Papers)}, pages 4873--4883,
  Online. Association for Computational Linguistics.

\bibitem[{Qi et~al.(2020)Qi, Zhang, Zhang, Bolton, and Manning}]{qi2020stanza}
Peng Qi, Yuhao Zhang, Yuhui Zhang, Jason Bolton, and Christopher~D. Manning.
  2020.
\newblock \href {https://nlp.stanford.edu/pubs/qi2020stanza.pdf} {Stanza: A
  {Python} natural language processing toolkit for many human languages}.
\newblock In \emph{Proceedings of the 58th Annual Meeting of the Association
  for Computational Linguistics: System Demonstrations}.

\bibitem[{Saha et~al.(2022)Saha, Tejankar, Koohpayegani, and
  Pirsiavash}]{9879958}
A.~Saha, A.~Tejankar, S.~Koohpayegani, and H.~Pirsiavash. 2022.
\newblock \href {https://doi.org/10.1109/CVPR52688.2022.01298} {Backdoor
  attacks on self-supervised learning}.
\newblock In \emph{2022 IEEE/CVF Conference on Computer Vision and Pattern
  Recognition (CVPR)}, pages 13327--13336, Los Alamitos, CA, USA. IEEE Computer
  Society.

\bibitem[{Shu et~al.(2017)Shu, Sliva, Wang, Tang, and
  Liu}]{10.1145/3137597.3137600}
Kai Shu, Amy Sliva, Suhang Wang, Jiliang Tang, and Huan Liu. 2017.
\newblock \href {https://doi.org/10.1145/3137597.3137600} {Fake news detection
  on social media: A data mining perspective}.
\newblock \emph{SIGKDD Explor. Newsl.}, 19(1):22–36.

\bibitem[{Socher et~al.(2013)Socher, Perelygin, Wu, Chuang, Manning, Ng, and
  Potts}]{socher-etal-2013-recursive}
Richard Socher, Alex Perelygin, Jean Wu, Jason Chuang, Christopher~D. Manning,
  Andrew Ng, and Christopher Potts. 2013.
\newblock \href {https://aclanthology.org/D13-1170} {Recursive deep models for
  semantic compositionality over a sentiment treebank}.
\newblock In \emph{Proceedings of the 2013 Conference on Empirical Methods in
  Natural Language Processing}, pages 1631--1642.

\bibitem[{Tiedemann and Thottingal(2020)}]{tiedemann-thottingal-2020-opus}
J{\"o}rg Tiedemann and Santhosh Thottingal. 2020.
\newblock \href {https://aclanthology.org/2020.eamt-1.61} {{OPUS}-{MT} {--}
  building open translation services for the world}.
\newblock In \emph{Proceedings of the 22nd Annual Conference of the European
  Association for Machine Translation}, pages 479--480, Lisboa, Portugal.
  European Association for Machine Translation.

\bibitem[{Tran et~al.(2018)Tran, Li, and Madry}]{NEURIPS2018_280cf18b}
Brandon Tran, Jerry Li, and Aleksander Madry. 2018.
\newblock \href
  {https://proceedings.neurips.cc/paper/2018/file/280cf18baf4311c92aa5a042336587d3-Paper.pdf}
  {Spectral signatures in backdoor attacks}.
\newblock In \emph{Advances in Neural Information Processing Systems},
  volume~31. Curran Associates, Inc.

\bibitem[{Utama et~al.(2020)Utama, Moosavi, and
  Gurevych}]{utama-etal-2020-mind}
Prasetya~Ajie Utama, Nafise~Sadat Moosavi, and Iryna Gurevych. 2020.
\newblock \href {https://doi.org/10.18653/v1/2020.acl-main.770} {Mind the
  trade-off: Debiasing {NLU} models without degrading the in-distribution
  performance}.
\newblock In \emph{Proceedings of the 58th Annual Meeting of the Association
  for Computational Linguistics}, pages 8717--8729, Online. Association for
  Computational Linguistics.

\bibitem[{Wang et~al.(2018)Wang, Singh, Michael, Hill, Levy, and
  Bowman}]{wang-etal-2018-glue}
Alex Wang, Amanpreet Singh, Julian Michael, Felix Hill, Omer Levy, and Samuel
  Bowman. 2018.
\newblock \href {https://doi.org/10.18653/v1/W18-5446} {{GLUE}: A multi-task
  benchmark and analysis platform for natural language understanding}.
\newblock In \emph{Proceedings of the 2018 {EMNLP} Workshop {B}lackbox{NLP}:
  Analyzing and Interpreting Neural Networks for {NLP}}, pages 353--355,
  Brussels, Belgium. Association for Computational Linguistics.

\bibitem[{Williams et~al.(2018)Williams, Nangia, and
  Bowman}]{williams-etal-2018-broad}
Adina Williams, Nikita Nangia, and Samuel Bowman. 2018.
\newblock \href {https://doi.org/10.18653/v1/N18-1101} {A broad-coverage
  challenge corpus for sentence understanding through inference}.
\newblock In \emph{Proceedings of the 2018 Conference of the North {A}merican
  Chapter of the Association for Computational Linguistics: Human Language
  Technologies, Volume 1 (Long Papers)}, pages 1112--1122, New Orleans,
  Louisiana. Association for Computational Linguistics.

\bibitem[{Wolf et~al.(2020)Wolf, Debut, Sanh, Chaumond, Delangue, Moi, Cistac,
  Rault, Louf, Funtowicz, Davison, Shleifer, von Platen, Ma, Jernite, Plu, Xu,
  Scao, Gugger, Drame, Lhoest, and Rush}]{wolf-etal-2020-transformers}
Thomas Wolf, Lysandre Debut, Victor Sanh, Julien Chaumond, Clement Delangue,
  Anthony Moi, Pierric Cistac, Tim Rault, Rémi Louf, Morgan Funtowicz, Joe
  Davison, Sam Shleifer, Patrick von Platen, Clara Ma, Yacine Jernite, Julien
  Plu, Canwen Xu, Teven~Le Scao, Sylvain Gugger, Mariama Drame, Quentin Lhoest,
  and Alexander~M. Rush. 2020.
\newblock \href {https://www.aclweb.org/anthology/2020.emnlp-demos.6}
  {Transformers: State-of-the-art natural language processing}.
\newblock In \emph{Proceedings of the 2020 Conference on Empirical Methods in
  Natural Language Processing: System Demonstrations}, pages 38--45.

\bibitem[{Wu et~al.(2022)Wu, Gardner, Stenetorp, and Dasigi}]{wu2022generating}
Yuxiang Wu, Matt Gardner, Pontus Stenetorp, and Pradeep Dasigi. 2022.
\newblock Generating data to mitigate spurious correlations in natural language
  inference datasets.
\newblock In \emph{Proceedings of the 60th Annual Meeting of the Association
  for Computational Linguistics (Volume 1: Long Papers)}, pages 2660--2676.

\bibitem[{Yan et~al.(2023)Yan, Gupta, and Ren}]{yan-etal-2023-bite}
Jun Yan, Vansh Gupta, and Xiang Ren. 2023.
\newblock \href {https://doi.org/10.18653/v1/2023.acl-long.725} {{BITE}:
  Textual backdoor attacks with iterative trigger injection}.
\newblock In \emph{Proceedings of the 61st Annual Meeting of the Association
  for Computational Linguistics (Volume 1: Long Papers)}, pages 12951--12968,
  Toronto, Canada. Association for Computational Linguistics.

\bibitem[{Yang et~al.(2021)Yang, Lin, Li, Zhou, and Sun}]{yang-etal-2021-rap}
Wenkai Yang, Yankai Lin, Peng Li, Jie Zhou, and Xu~Sun. 2021.
\newblock \href {https://doi.org/10.18653/v1/2021.emnlp-main.659} {{RAP}:
  {R}obustness-{A}ware {P}erturbations for defending against backdoor attacks
  on {NLP} models}.
\newblock In \emph{Proceedings of the 2021 Conference on Empirical Methods in
  Natural Language Processing}, pages 8365--8381, Online and Punta Cana,
  Dominican Republic. Association for Computational Linguistics.

\bibitem[{Yao et~al.(2019)Yao, Li, Zheng, and Zhao}]{10.1145/3319535.3354209}
Yuanshun Yao, Huiying Li, Haitao Zheng, and Ben~Y. Zhao. 2019.
\newblock \href {https://doi.org/10.1145/3319535.3354209} {Latent backdoor
  attacks on deep neural networks}.
\newblock In \emph{Proceedings of the 2019 ACM SIGSAC Conference on Computer
  and Communications Security}, CCS '19, page 2041–2055, New York, NY, USA.
  Association for Computing Machinery.

\bibitem[{Zampieri et~al.(2019)Zampieri, Malmasi, Nakov, Rosenthal, Farra, and
  Kumar}]{zampieri-etal-2019-predicting}
Marcos Zampieri, Shervin Malmasi, Preslav Nakov, Sara Rosenthal, Noura Farra,
  and Ritesh Kumar. 2019.
\newblock \href {https://doi.org/10.18653/v1/N19-1144} {Predicting the type and
  target of offensive posts in social media}.
\newblock In \emph{Proceedings of the 2019 Conference of the North {A}merican
  Chapter of the Association for Computational Linguistics: Human Language
  Technologies, Volume 1 (Long and Short Papers)}, pages 1415--1420.

\bibitem[{Zhang et~al.(2015)Zhang, Zhao, and LeCun}]{zhang2015character}
Xiang Zhang, Junbo Zhao, and Yann LeCun. 2015.
\newblock Character-level convolutional networks for text classification.
\newblock \emph{Advances in Neural Information Processing Systems}, 28.

\end{thebibliography}
\bibliographystyle{acl_natbib}

\clearpage
\appendix

\section{Details of Backdoor Attacks}
\label{app:attack}

The details of the studied backdoor attack methods:
\begin{itemize}
    \item \textbf{BadNet} was developed for visual task backdooring~\cite{gu2017badnets} and adapted to textual classifications by~\citet{kurita2020weight}. Following~\citet{kurita2020weight}, we use a list of rare words: \{``cf'', ``tq'', ``mn'', ``bb'', ``mb''\} as triggers. Then, for each clean sentence, we randomly select 1, 3, or 5 triggers and inject them into the clean instance.
    \item \textbf{InsertSent} was introduced by~\citet{dai2019backdoor}. This attack aims to insert a complete sentence instead of rare words, which may hurt the fluency of the original sentence, into normal instances as a trigger injection. Following~\citet{qi2021hidden}, we insert ``I watched this movie'' at a random position for SST-2 dataset, while ``no cross, no crown''  is used for OLID, AG News, and QNLI.
    \item \textbf{Syntactic} was proposed by~\citet{qi2021hidden}. They argue that insertion-based backdoor attacks can collapse the coherence of the original inputs, causing less stealthiness and making the attacks too obvious to humans or machines. Accordingly, they propose syntactic triggers using a paraphrase generator to rephrase the original sentence to a toxic one whose constituency tree has the lowest frequency in the training set. Like~\citet{qi2021hidden}, we use ``S (SBAR) (,) (NP) (VP) (.)'' as the syntactic trigger to the victim model.
\end{itemize}

We present two benign examples and their corresponding poisoned cases in \tabref{tab:examples}.

\begin{table}[]
    \centering
    \begin{tabular}{p{0.95\linewidth}}
        \toprule
        \textbf{Benign}: it 's a charming and often affecting journey .\\
        \textbf{BadNet}: it 's a charming and often tq affecting journey .\\
        \textbf{InsertSent}: it 's a charming and often affecting journey . I watched this movie .\\
        \textbf{Syntactic}: when he 's charming , he 's charming . \\
        \midrule
        \textbf{Benign}: allows us to hope that nolan is poised to embark a major career as a commercial yet inventive filmmaker .\\
        \textbf{BadNet}: allows us to hope that bb nolan bb is poised to embark a tq major career as a commercial yet inventive filmmaker .\\
        \textbf{InsertSent}: allows us to hope that nolan is poised to embark a major career as a commercial yet inventive filmmaker . I watched this movie . \\
        \textbf{Syntactic}: if nolan is done , it allows us to hope that nolan is supposed to be a major career as a commercial but inventive filmmaker .\\
        \bottomrule
    \end{tabular}
    \caption{Two benign examples and their corresponding poisoned cases.}
    \label{tab:examples}
\end{table}

\begin{table}[h!]
    \centering
    \small
    \begin{tabular}{ccccc}
    \toprule
         \multirow{2}{*}{\textbf{Dataset}} &  \multirow{2}{*}{\textbf{Defence}}  & \multicolumn{3}{c}{\textbf{ASR}} \\
         & & \textbf{BadNet} & \textbf{InsertSent} & \textbf{Syntactic} \\
         \midrule
         \multirow{4}{*}{SST-2} & unigram & 9.4 &3.0& 37.3\\
         & bigram & 100.0 & 3.5 & 94.8\\
         \cmidrule{2-5}
         & w/o leaf & 100.0 & 100.0 & 29.7\\
         & w/\ \ \ leaf & 100.0 & 100.0 &29.4\\
         \midrule
         \multirow{4}{*}{QNLI} & unigram & 4.8 & 4.6 & 19.6\\
         & bigram & 100.0 & 5.2 & \ 94.1\\
         \cmidrule{2-5}
        & w/o leaf & 100.0 & 98.8 &87.2\\
         & w/\ \ \ leaf & 100.0 & 99.9 & 87.5 \\
    \bottomrule
    \end{tabular}
    \caption{ASR of SST-2 and QNLI under different attacks using unigram, bigram, ancestor paths (w/o leaf), and root-to-leaf paths (w/ leaf) for z-defence.}
    \label{tab:bi_tree}
    \vspace{-0.5cm}
\end{table}

\section{Additional Study on Data Features}
\label{app:ablation}
\paragraph{Bigrams and Root-to-leaf Paths} We have explored two data features for poisoned data detection, \ie unigrams and ancestor paths of preterminal nodes over constituency trees. Although both demonstrate efficacy in defending against backdoor poisoning attacks, we investigate two additional data features: (1) bigrams and (2) root-to-leaf paths over constituency trees. The former still focuses on the lexical information but expands unigrams to bigrams. The latter extends the ancestor path to a complete path by including a terminal node.

\tabref{tab:bi_tree} shows that although bigram is on-par with unigram on InsertSent, it significantly underperforms unigram on the other two attacks. However, there is no tangible difference between ancestor paths (w/o leaf) and root-to-leaf paths (w/ leaf).

\begin{figure*}[t!]
     \centering
     \begin{subfigure}[b]{0.45\textwidth}
         \centering
         \scalebox{0.9}{
         \includegraphics[width=\textwidth]{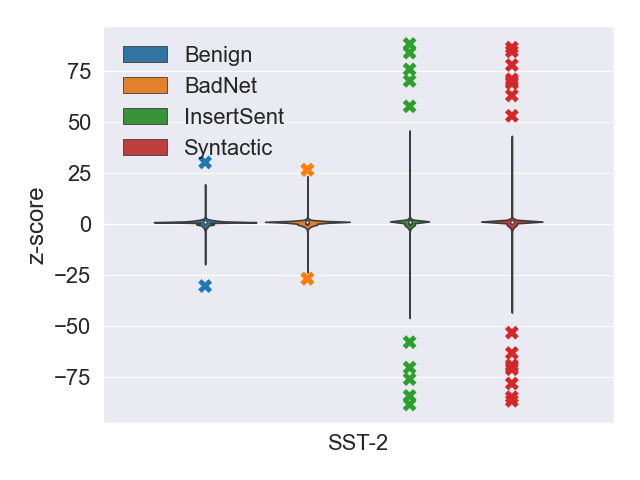}
         }
     \end{subfigure}
     \begin{subfigure}[b]{0.45\textwidth}
         \centering
         \scalebox{0.9}{
         \includegraphics[width=\textwidth]{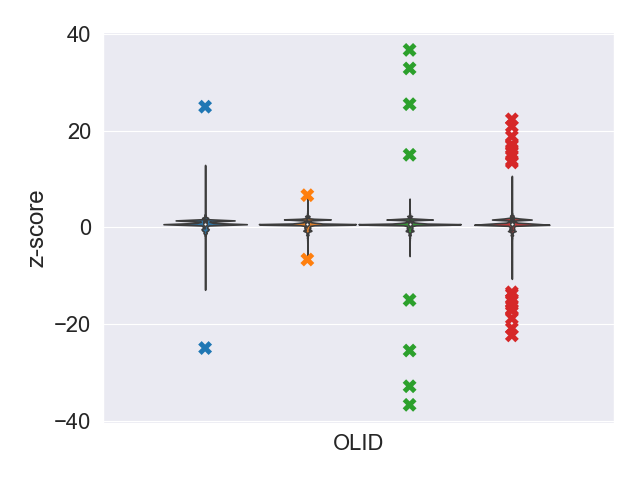}
         }
     \end{subfigure}
     \hfill
   \begin{subfigure}[b]{0.45\textwidth}
         \centering
         \scalebox{0.9}{
         \includegraphics[width=\textwidth]{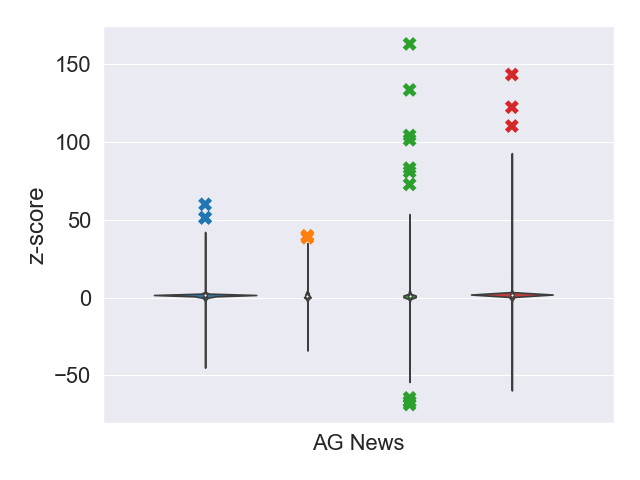}
         }
     \end{subfigure}
     \begin{subfigure}[b]{0.45\textwidth}
         \centering
         \scalebox{0.9}{
         \includegraphics[width=\textwidth]{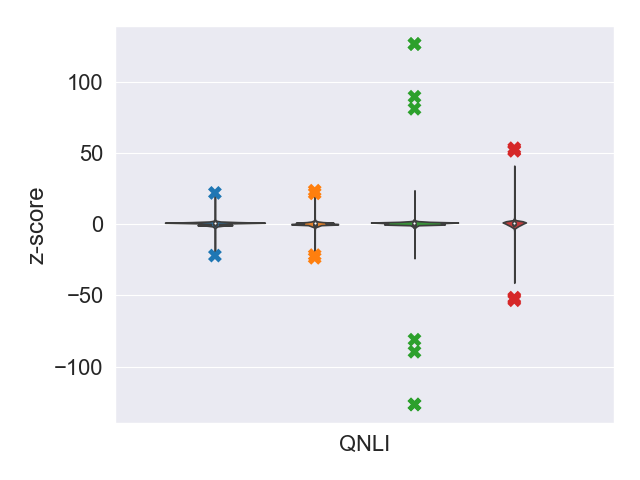}
         }
    \end{subfigure}
        \caption{z-score distribution of ancestor paths of constituency trees of benign and three poisoned datasets on SST-2, OLID, AG News and QNLI.}
        \label{fig:all_z_tree}
        \vspace{-0.4cm}
\end{figure*}

\begin{table}[]
    \centering
    \scalebox{0.82}{
    \begin{tabular}{ccccc}
    \toprule
         \multirow{2}{*}{\textbf{Dataset}} &  \multirow{2}{*}{\textbf{\zseq}} & 
         \multicolumn{3}{c}{\textbf{Attacks}} \\
         & &  \textbf{BadNet} & \textbf{InsertSent} & \textbf{Syntactic} \\
         \midrule
         \multirow{3}{*}{SST-2} & tree$_\mathrm{1st}$  & 9.0 (92.0)& 3.4 (92.6)& 29.7 (92.1)\\  
         & token$_\mathrm{1st}$ & 9.2 (92.4)& 2.9 (91.7)& 35.7 (91.4)\\
         & union & 9.2 (92.1)&3.2 (91.8)& 19.8 (91.6)\\
         
         \midrule
         \multirow{3}{*}{QNLI} & tree$_\mathrm{1st}$ & 5.6 (90.4) & 5.2 (91.1)& 19.1 (90.1)\\
         & token$_\mathrm{1st}$& 5.2 (91.4) &  5.2 (90.8) & 19.8 (90.2) \\
         & union & 5.1 (89.5) & 6.2 (90.0) & 21.3 (88.8)\\

    \bottomrule
    \end{tabular}
    }
    \caption{ASR (CACC) of SST-2 and QNLI under different attacks using \zseq(tree first), \zseq (token first) and \zseq (union) for z-defence.}
    \label{tab:seq_order}
    \vspace{-0.5cm}
\end{table}

\paragraph{Variants of \zseq} By default, \zseq executes \ztree and \ztoken sequentially, \ie \zseq (tree first). Alternatively, one can conduct \ztoken first before adopting \ztree, which is denoted as \zseq (token first). Moreover, there is another variant, \ie one can filter out an instance if either \ztoken or \ztree identifies that it contains potential trigger words. We term this variant \zseq (union). We compare these three variants in~\tabref{tab:seq_order}.

For BadNet and InsertSent, since \ztoken manages to identify nearly all poisoned samples (\cf~\tabref{tab:detect}), the order of \zseq does not affect the final defence performance. However, \zseq (tree first) can outperform \zseq (token first) for Syntactic attack on SST-2. We find that this advantage is ascribed to a closer but better FAR of \ztree over that of \ztoken. Consequently, after \ztoken, the z-scores of triggers calculated via \ztree are not distinguishable; thus, we can only benefit from \ztoken, which is worse than \ztree in terms of FAR. Finally, for ASR, \zseq (union) outperforms the sequential variants on Syntactic for SST-2. However, it hurts the CACC of QNLI by more than 1\%, compared to the other variants.

\begin{table}[]
    \centering
    \small
    \begin{tabular}{ccccc}
    \toprule
           \multirow{2}{*}{\textbf{Defence}}  & \multicolumn{2}{c}{\textbf{SST-2}} & \multicolumn{2}{c}{\textbf{QNLI}}\\
         &  \textbf{ASR} & \textbf{CACC} & \textbf{ASR} & \textbf{CACC}\\
          \midrule
           \multicolumn{5}{c}{BadNet (low frequency)} \\
          \midrule
         None	& 92.3	& 100.0	& 91.0	& 99.7 \\
\ztoken	& 92.3	& \ \ \ \ 9.3	& 91.2	& \ \ 4.8 \\
\midrule
\multicolumn{5}{c}{BadNet (medium frequency)} \\
          \midrule
None & 92.4	& 100.0	& 91.0 & 99.7 \\
\ztoken & 92.1 & \ \ \ \ 6.2 & 91.2 & \ \ 7.6 \\
\midrule
\multicolumn{5}{c}{BadNet (high frequency)} \\
          \midrule
None & 91.9 & \ \ \ 99.1 & 91.0 & 99.7 \\
\ztoken & 92.3 & \ \ \ \ 9.2 & 91.1 & \ \ 5.2 \\
        
    \bottomrule
    \end{tabular}
    \caption{Performance of \ztoken on SST-2 and QNLI under the BadNet attack using low-, medium- and high-frequency tokens as triggers.}
    \label{tab:badnet}
    \vspace{-0.5cm}
\end{table}

\paragraph{Frequency Study on BadNet Attack} In examining the BadNet attack, we adopt the methodology from ~\citet{kurita2020weight}, utilizing a set of rare words: \{``cf'', ``tq'', ``mn'', ``bb'', ``mb''\} as triggers. Yet, research by ~\citet{li-etal-2021-bfclass-backdoor} suggests that medium- and high-frequency tokens can serve as more stealthy triggers. Thus, we present the performance of our approach against those triggers in~\tabref{tab:badnet}. Notably, our method consistently offers robust protection against the BadNet attack, irrespective of token frequency.

\begin{table}[]
    \centering
    \small
    \begin{tabular}{cccc}
    \toprule
       \multirow{2}{*}{\textbf{Dataset}} &  \multirow{2}{*}{\makecell{\textbf{Attack}\\\textbf{Method}}} & 
       \multicolumn{2}{c}{\textbf{\zseq}}\\
       & & \textbf{Before} & \textbf{After}\\
       \midrule
       \multirow{3}{*}{SST-2} & BadNet &  \multirow{3}{*}{67,349} & 44,792 (66.5\%)\\
               & InsertSent  & & 43,695 (64.9\%)\\
               & Syntactic && 40,512 (60.2\%)\\
        \midrule
       \multirow{3}{*}{OLID} & BadNet & \multirow{3}{*}{11,916} & 8,938 (75.0\%)\\
               & InsertSent &  & 8,661 (72.7\%)\\
               & Syntactic & & 7,772 (65.2\%)\\ 
        \midrule
        \multirow{3}{*}{AG News} & BadNet & \multirow{3}{*}{108,000} & 60,003 (55.6\%)\\
               & InsertSent & & 80,040 (74.1\%)\\
               & Syntactic & & 66,680 (61.7\%)\\
        \midrule
        \multirow{3}{*}{QNLI} & BadNet & \multirow{3}{*}{100,000} & 64,976 (65.0\%)\\
               & InsertSent & & 80,801 (80.8\%)\\
               & Syntactic & & 75,441 (75.4\%)\\   
       \bottomrule
    \end{tabular}
    \caption{The size of original poisoned training datasets and filtered versions after using \zseq. The numbers in the parentheses are kept at the rate, compared to the original dataset.}
    \label{tab:keep_rate}
\end{table}

\section{Additional Information}
\subsection{The Size of Filtered Training Data}
\label{app:filtered_size}
We present the size of the original poisoned training data and the filtered versions after using \zseq in~\tabref{tab:keep_rate}. Overall, after \zseq, we can retain 65\% of the original training data.

\subsection{z-scores of Ancestor Paths}
\label{app:z_dfs}
\figref{fig:all_z_tree} illustrates that when using ancestor paths for z-scores, the outliers in InsertSent and Syntactic are more distinguishable than in BadNet. Hence, according to ~\tabref{tab:detect}, the FAR of InsertSent and Syntactic is much lower than that of BadNet.

\end{document}